\documentclass[a4paper, 10pt, conference]{ieeeconf}      

\IEEEoverridecommandlockouts                              

\usepackage{amsmath}
\usepackage{graphicx}

\usepackage{float}
\usepackage{multicol}
\usepackage{multirow}
\usepackage{epstopdf}
\usepackage{enumerate}
\usepackage[dvipsnames]{xcolor}
\usepackage{comment}
\usepackage[inkscapelatex=false]{svg}
\usepackage{pdfpages}

\title{\LARGE \bf
Integration of Vision-based Object Detection and Grasping for Articulated Manipulator in Lunar Conditions
}

\author{
Camille Boucher$^{*1,2}$, Gustavo H. Diaz$^{2}$, Shreya Santra$^{2}$, Kentaro Uno$^{2}$, and Kazuya Yoshida$^{2}$
\thanks{*This work is part of the Moonshot Goal 3 supported by the Japan Science and Technology Agency}
\thanks{$^{1}$C. Boucher is with IMT Atlantique, Brest 29238, France
        {\tt\small camille.boucher@imt-atlantique.net}}%
\thanks{$^{2}$G.H. Diaz, S. Santra, K. Uno, and K. Yoshida are with Space Robotics Lab. in the Department of Aerospace Engineering, Graduate School of Engineering, Tohoku University, 
        Sendai 980--8579, Japan
        {\tt\small diaz.huenupan.gustavo.hernan.p3@dc.tohoku.ac.jp}}%
}%

\newcommand{\fig}[1]{Fig. \ref{#1}}
\newcommand{\tab}[1]{Table \ref{#1}}

\begin{document}

\maketitle
\thispagestyle{empty}
\pagestyle{empty}

\begin{abstract}
The integration of vision-based frameworks to achieve lunar robot applications faces numerous challenges such as terrain configuration or extreme lighting conditions. This paper presents a generic task pipeline using object detection, instance segmentation and grasp detection, that can be used for various applications by using the results of these vision-based systems in a different way. We achieve a rock stacking task on a non-flat surface in difficult lighting conditions with a very good success rate of 92\%. Eventually, we present an experiment to assemble 3D printed robot components to initiate more complex tasks in the future.

\end{abstract}

\section{Introduction}
The ability to observe and understand the environment has been expanded to robotic systems with artificial intelligence and machine learning has demonstrated its use to achieve impressive outcomes in various fields, including image and data processing for robotic application. 
Imitating the human ability to detect and grasp any sort of object has posed challenges for applications such as transporting large objects, construction and automation. The combination of machine vision and robotics to replicate such type of grasping requires precise target detection, localization and manipulation.

\begin{figure}[t]
      \centering
      \includegraphics[scale=0.56]{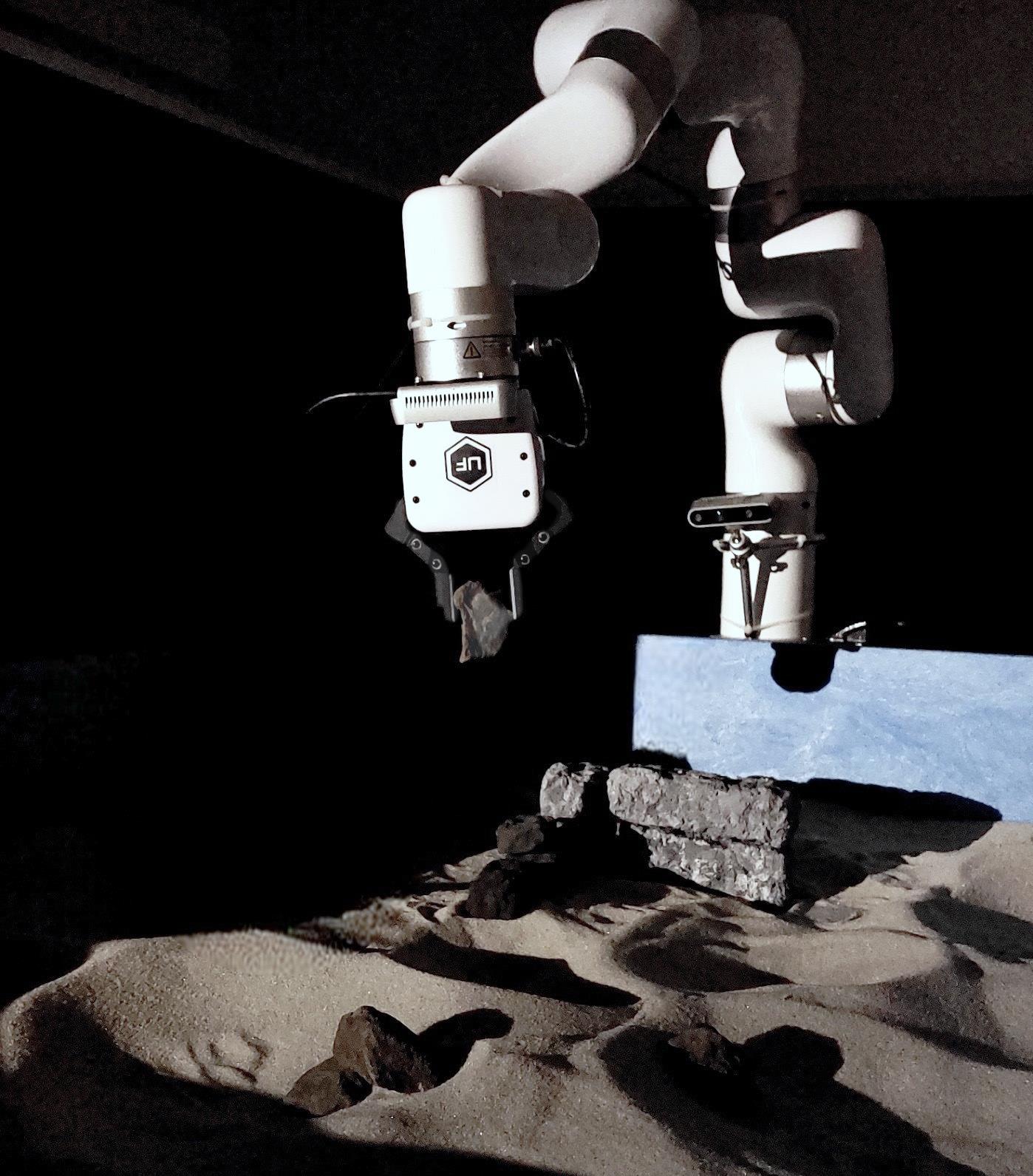}
      \caption{Robot-xArm7, gripper with camera and second camera fixed at the base, stacking rocks in a moon-like environment.}
      \label{moon}
\end{figure}

This paper aims to tackle this challenge in lunar conditions with limited lighting conditions, considering various craters, environment changes and irregular objects like in \fig{moon}. A lot of unpredictable situations can occur in a lunar mission without any possibility of human assistance. The robots must achieve missions of exploration, scientific experiments, construction, etc., using accurate and robust neural networks. We aim to demonstrate that a generic software architecture using the vision-based neural networks YOLOv8 (You Only Look Once) \cite{YOLO8} and GPD (Grasp Pose Detection) \cite{Pas17}, as shown in \fig{generic_pipeline}, can be used to achieve numerous applications like rock stacking or autonomous robot assembling, and this by just modifying the dataset and the manipulation pipeline. The goal is, therefore, not to improve the existing neural networks but to integrate and use them efficiently to perform the aforementioned tasks.


We will first establish state-of-the-art on the different elements of the generic pipeline: object detection, instance segmentation and grasp detection.
A brief overview of our system used for this paper will be done in the second part.
Then, we will present our vision-based frameworks: YOLOv8 used on custom datasets and GPD. 
Subsequently, we will explain the ROS integration, and two applications - rock stacking and robot assembling - by just changing the way of using the results from the vision-based system.
Eventually, our experiments will be presented, followed by an analysis and the outline for the future.

\section{state of the art }

\subsection{Object Detection}



In visual object recognition, the use of Convolutional Neural Network (CNN) has led to new challenges. The detectors can be classified into two categories: two-stage or regional-proposal-based algorithms and single-stage ones. One-stage frameworks have the advantage to process the entire image in a single pass, making it more computationally efficient and better suited for real-time detection \cite{Liu19}.

In 2015, J. Redmon \emph{et al.} presented a new one-shot framework YOLO \cite{Red16}. 
J.Terven \emph{et al.} analyzed the YOLO’s evolution, examining the innovations and contributions in each iteration from the original YOLO to the new version YOLOv8 in January 2023 \cite{Ter23}. The first version performed faster than any existing object detector but the localization error was larger compared with state-of-the-art methods such as region-based Fast R-CNN \cite{Gir15}. Through the years, YOLO has evolved to stand out as state-of-the-art object detection in a real-time framework with its remarkable balance of speed and accuracy. It has then been used in numerous fields such as autonomous vehicles with object tracking, like pedestrians \cite{Lan18} and other obstacles \cite{Daz22}, surveillance and security fields \cite{Ash22} or medical field with cancer detection \cite{Nie19}. D. Reis \emph{et al.} demonstrated the use of the latest version YOLOv8 for detecting flying objects in real time in a challenging environment \cite{Rei23}.

\subsection{Instance Segmentation}
Along with the object detection challenge, the semantic segmentation and the instance segmentation are also very discussed problems. While object detection aims to classify and give the location, the goal of semantic segmentation is to label every pixel into a class according to the region within which it is enclosed. 
A.M. Hafiz \emph{et al.} defined the instance segmentation problem as the task of providing simultaneous solutions to object detection as well as semantic segmentation \cite{Haf20}. They reviewed in 2020 the evolution of instance segmentation up to Mask R‑CNN \cite{He18}, YOLACT \cite{Bol19} and TensorMask \cite{Chen19}. As for the object detection the one-shot models are said to be faster than the two-stage ones, and therefore more suitable for real time utilization. 

\subsection{Grasping Detection}
To allow robots to achieve various tasks and reproduce human behaviours, the challenge of reliably grasping and handling objects, like household items, mechanical parts or packages, is extremely important.
The research on robotic systems for manipulation tasks has mainly focused on human-robot interaction, and at first, systems were lacking in the autonomous part of grasping and placing an unknown object in an unstructured environment.
Mahler \emph{et al.} proposed DexNet, a grasp system, with a 93\% grasp success rate,  which takes depth images as input and gives grasps in the plane as output, i.e. with a single degree of orientation freedom around the gravity axis \cite{Mah17}.
Morrison \emph{et al.} \cite{Mor18} and Viereck \emph{et al.} \cite{Vie17} studied the problem of grasping dynamically moving objects and proposed a closed loop system with a grasp success rate of 83\% and 88.9\%.
Gualtieri \emph{et al.} proposed GPD framework \cite{Gua16}\cite{Pas17} that takes point cloud data as input and produces 6-DOF grasp poses as output. Their system incorporates a new method for generating grasp hypotheses that, relative to prior methods, does not require a precise segmentation of the object to be grasped and can generate hypotheses on any visible surface.
Their system gives really good results, especially for dense environments with a grasp success rate of 89\%. In the final step of their work, they also discussed the idea of combining object and grasp detection. They made experiments on household objects, but only evaluating the accuracy of the object detection on the grasped objects, the grasping strategy was not based on the object detection results such as proposed here.

\section{System Overview}
Our robotic system is comprised of an articulated arm xArm7 (7-DOF) from UFactory.
It is equipped with a parallel gripper with two fingers; the robotic arm is fixed on a table next to the sandbox. The vision system is made up of two Intel RealSense cameras d435 which retrieve RGB-D (color and depth image). 
To recreate lunar-like conditions with uneven surfaces we use sand and an artificial light source as shown in \fig{moon}.
For the manipulations we use various irregular objects such as polystyrene rocks and 3D printed robot components like head, body, joint, etc. \fig{custom_datasets}. These objects are challenging because of their irregularities in shape, size, color and weight.\\ 
The computer used is equipped with CPU Intel 19.13900KF 24 cores and GPU NVIDIA GeForce RTX 4090/24 GB. 

\begin{figure}[b]
      \centering
      \includegraphics[scale=0.28]{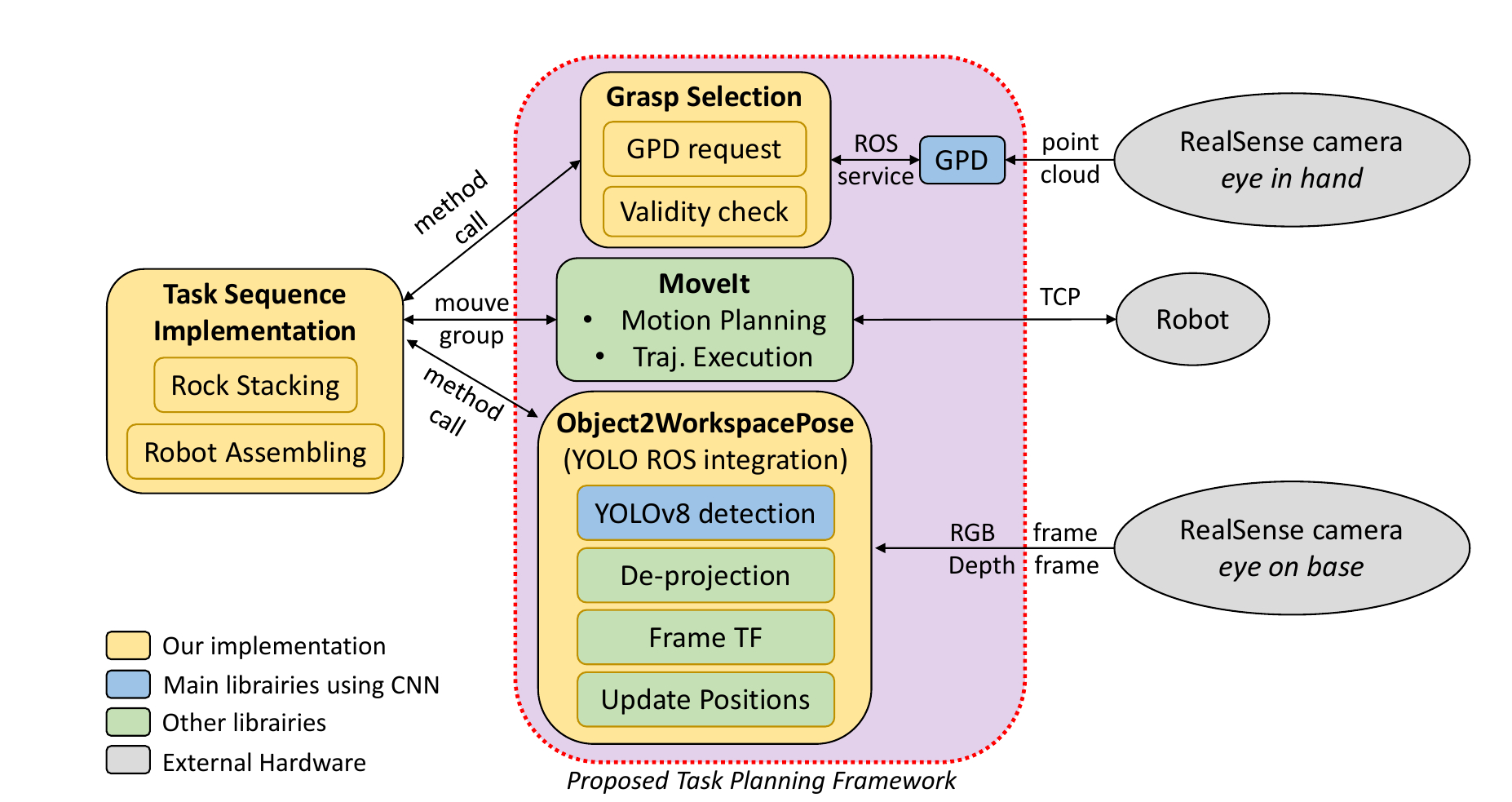}
      \caption{Generic software architecture based on vision-based frameworks.}
      \label{generic_pipeline}
   \end{figure}

The software system shown in \fig{generic_pipeline} is comprised of some low-level and medium-level packages, like MoveIt, for the controls, motion planning, etc. and a high level with various applications like object detection or robot assembling. 

\section{Vision-based frameworks }

\subsection{YOLOv8}

To perform object detection and instance segmentation, we choose YOLOv8 \cite{YOLO8}, whose new architecture is well resumed by J.Terven \emph{et al.} \cite{Ter23}. Solawetz \emph{et al.}  explained the improvements from the previous versions such as the anchor free detection and the new convolutions \cite{Sol23}.
This version has a high mean Average Precision (mAP) (respectively 50.2 and 53.9 mAP50-95 for its medium model and larger one) while maintaining a lower inference speed on the COCO (Common Objects in Context) dataset \cite{COCO}. Another positive highlight is that YOLOv8 can be used both with a command line interface and with a PIP package, which is very useful for ROS integration and for all the tasks like training, validation, prediction, etc.

\subsection{Custom dataset and YOLOv8 training }

In order for our object detection results to be applicable in lunar robotic applications, it needs to perform efficiently in a challenging environment with shadows, high exposure, occlusion, and on miscellaneous objects such as robot parts, screws, bolts, various types of rocks, etc. The construction and the training of a custom dataset, considering the identified complexities, are as important as the model choice to achieve highly accurate results.

\begin{figure}[t]
      \centering
      \includegraphics[scale=0.32]{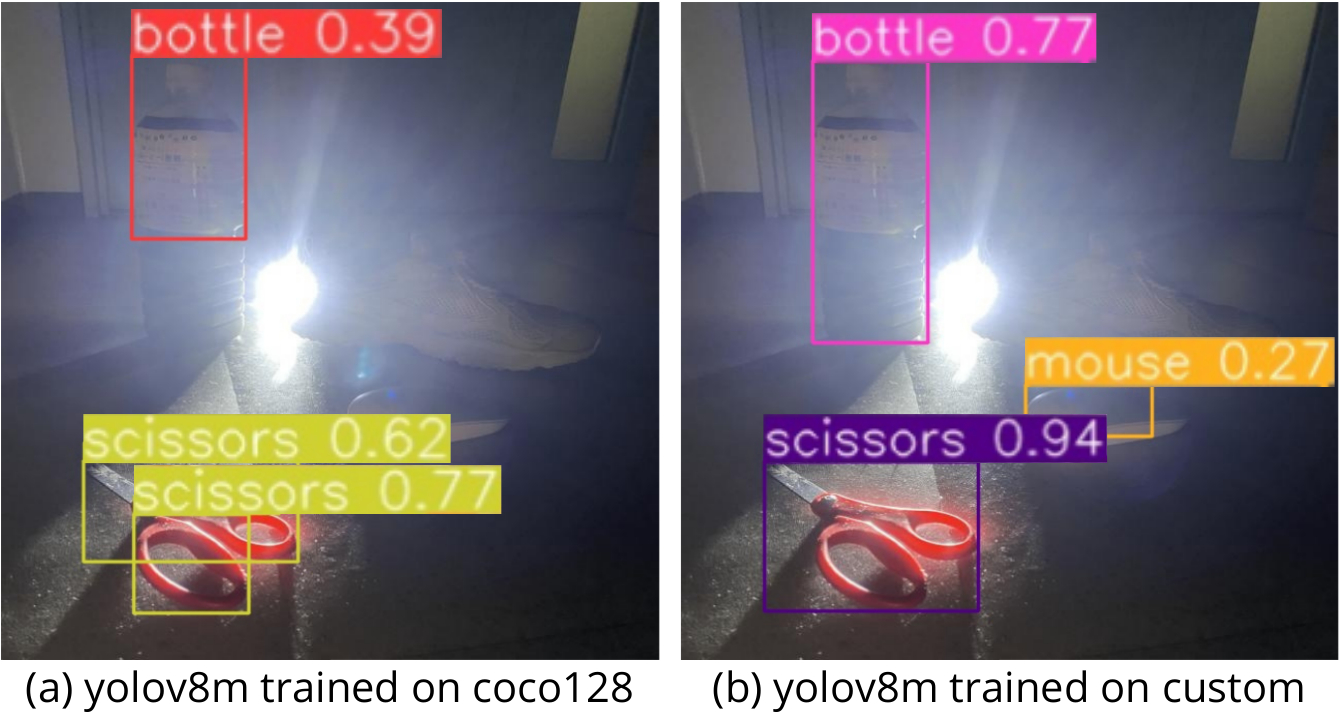}
      \caption{Object detection on bottles, mouses and scissors in difficult lighting.}
      \label{coco128}
   \end{figure}

To better highlight the importance of a custom dataset, especially considering the lighting in a lunar scenario, we compare the mean Average Precision between two models. The first one is YOLOv8m, YOLOv8 medium model, trained on coco128 \cite{coco128}, a sub-dataset of 128 images from the COCO dataset. The second is YOLOv8m trained on a custom dataset. We add to the coco128 images 62 new pictures of 6 objects (bottle, laptop, mouse, scissors, spoon and fork) in more complex lighting conditions than in the original dataset. Examples images from these datasets are shown in \fig{coco128} where three objects are shown (scissor, mouse and bottle). For the validation, we ensure to use different objects than the ones used for the training and different lighting conditions. We also make different validation sets by adding several augmentation steps which degrade image quality. We can see in Table 1 that even with the more complex validation set (brightness, exposure and 10\% of noise) the model trained with the custom dataset outperforms the original one. Therefore, during the construction of our datasets, we take particular care to include a wide variety of images in different lighting and exposure conditions, occluded and cropped objects, different colors, sizes and shapes, etc.

\begin{table}[b]
\caption{Object detection accuracy of YOLOv8 models trained with different datasets, on different validation sets.}
\label{table_example} 
\begin{center}
\begin{tabular}{|c||c||c||c|}
\hline
Training dataset & Validation set & mAP50  & mAP50-95\\
\hline
COCO128 & normal & 23.3 & 22.9\\
\hline
Custom & b/e* & 61.4 & 52.3\\
\hline
Custom &  b/e* - noise 5\% & 40.4 & 31.0\\
\hline
Custom &  b/e* - noise 10\% & 36.3 & 28.1\\
\hline
\end{tabular}
\end{center}
* augmentation step on the validation set: brightness and exposure +/- 25\%

\end{table}

\begin{figure}[b]
      \centering
      \includegraphics[scale=0.4]{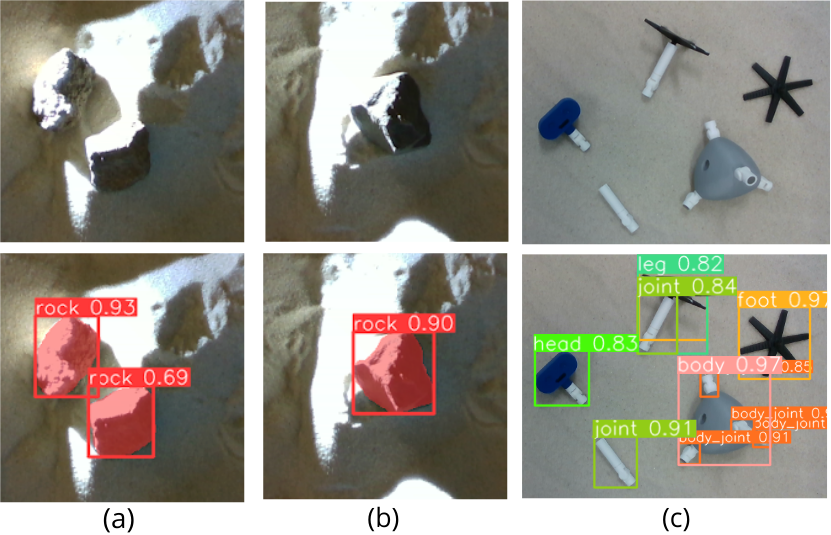}
      \caption{Custom datasets and YOLOv8 predictions (a) moonrock segmentation (b) moonrock segmentation, mask deformed by strong exposure (c) robot components detection.}
      \label{custom_datasets}
   \end{figure}

For the different applications, we constructed two main datasets, each with different challenges.
First, polystyrene-made imitated rocks with the main challenges being lighting conditions and uneven surfaces, \fig{custom_datasets} (a) and (b). During the dataset creation, the sand,  the lighting and the exposure conditions were modified. We also add different augmentation steps: +/- 25\% of brightness and exposure and 5\% of noise. The augmentation enables  the dataset to be artificially enlarged using label-preserving transformation to reduce over-fitting on image data. We include real rocks in the validation to get more reliable accuracy results. Instance segmentation will be performed on this dataset, the validation results will be presented in part VI.

The second dataset is 3D printed robot components - head, body, joint, legs, etc. \fig{custom_datasets} (c), with two main challenges. The first one is low inter-class variance: components which look very similar to each other compared to the rest of the labels, for example, the difference between a joint and a body joint consists in being attached or not to a robot body. The second challenge has overlapping classes and occlusions. Using the YOLOv8 results we are able to recognize the robots components and determine their state: if the algorithm detects a body, it will need to get its associated joints, furthermore if a body joint is detected it needs to be classified as available or not (another part already attached to or not), and likewise. 
The labelling rules are very important and need to be defined before the annotations; in this dataset , for example, we decide to define a leg as a joint plus a foot and a body and the main body part plus its body joints as we can see in \fig{custom_datasets}. 

After creating the datasets we train them on YOLOv8 models. The robot dataset is composed of 528 images with 12 classes; which are split into 90\% as the training set and 10\% as the validation set. We train on different epochs to detect the moment where the model stops improving and begins overfitting. For this model it is around 50 epochs (about 3 times the number of classes). We also decide to keep the original training hyper-parameters, since the dataset is not very large, we do not want to over-fit the model by tuning the parameters. The hyper-parameters are: batch size of 16, AdamW as the optimizer, momentum of 0.937, weight decay of 0.0005 and learning rate of 0.000667.

We perform the training of the small, medium and large YOLOv8 models and then evaluate to determine an optimal trade-off between inference speed and mAP50-95.
We can see in Table II a noticeable mAP50-95 increase between the small and the medium model but not a significant improvement between the medium and the large. On the validation set all the models have an average total speed (pre-process + inference + post-process) under 10 milliseconds/image, which fit perfectly with the detection in real time.  Regarding the results, we decide to choose the medium model YOLOv8m. We test on different SDR videos and we observe a total speed of 0.71 + 7.34 + 0.89 = 8.94 milliseconds, more than 60 fps (frames per second) which is consistent with real-time use.

\begin{table}[b]
\caption{YOLOv8 models training on robot dataset results.}
\label{robot_model_training_table} 
\begin{center}
\begin{tabular}{|c||c||c||c||c|}
\hline
Model & Parameters & Layers & mAP50-95 & Speed (ms)\\
\hline
Small & 11139857 & 225 & 81.9 & 1.8\\
\hline
Medium & 25862689 &295 & 84.3 & 3.4\\
\hline
Large & 43638321 & 365 & 84.5 & 8.5\\
\hline
\end{tabular}
\end{center}

\end{table}

\subsection{GPD configuration and tuning}
For the object manipulations, the results obtained with YOLOv8 are not enough and we need to improve the grasping strategy.
We integrate GPD package for the grasp detection for several reasons; firstly, because it can be easily integrated to ROS with a package in C++ and Python. Moreover, since GPD operates with point cloud input, we can easily manipulate this point cloud using the results obtained from object detection. In addition, since GPD is not limited to detecting planar grasps, it can more easily generate side grasps, which can be needed for some rocks or robots components, it then better ensures the autonomous grasping in any kind of situation.

The GPD library allows configuring several parameters related to i) the geometry of the gripper and grasp descriptor, ii) pre-processing of the point cloud, iii) grasp candidates generation, iv) filters and selection.\\
For the hand geometry, we first test with the actual dimensions of our UFactory gripper, however, we find that we can get more successful grasps using smaller dimensions, according to the size of the objects.
This also reduces the computation time, since the grasp descriptor is based on the volume inside the gripper.\\
For ii), we set the workspace parameters to match the field of view of the point cloud from the \textit{pre-grasp position} in order to generate candidates only around the observed object of interest. It is also necessary to set the \textit{sample\_above\_plane} to filter out candidates on the table plane.\\
For iii), the first important parameter is the \textit{hand\_axes} to define the main axis of the generated candidates. We select a vertical orientation that facilitates the actual planning and execution trajectory. Second is the \textit{number of orientation} and \textit{number of samples} generated around the selected axis, we find that 5 orientations and 100 samples are sufficient to find valid grasp candidates in real-time.\\
For iv), we enable the \textit{filter by approach direction} in the z axis, again to facilitate the planning and execution; setting the number of selected grasps to 20 is also enough to ensure a real-time selection of valid candidates.

\section{Integration on xArm7}

\subsection{Rock stacking in a moon-like environment}

\begin{figure}[b]
      \centering
      \includegraphics[scale=0.28]{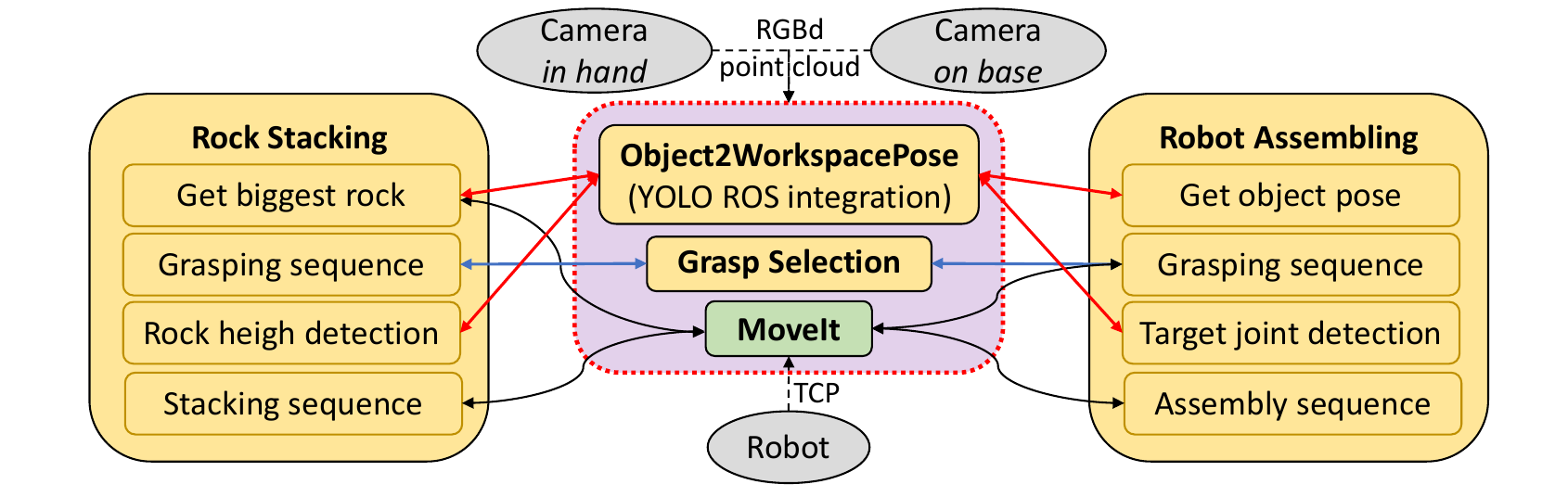}
      \caption{Rock stacking and Assembly tasks vision-based frameworks.}
      \label{software_architecture}
\end{figure}

In this part, we will focus on the use of the vision-based frameworks results for the stacking rocks task. For lunar exploration, we want robots being able to autonomously recognize interesting rocks, pick and place them for analysis. We also want the robots to achieve construction tasks. Therefore, our first application is to autonomously stack small and medium rocks in these lunar conditions.
We use our vision-based general framework shown in \fig{generic_pipeline} for specific sub-tasks as described in \fig{software_architecture}.

The first step is to classify all the detected rocks by size to stack them in decreasing order; the sorting is done using the area of the object's mask given by the instance segmentation.

\begin{figure}[t]
      \centering
      \includegraphics[scale=0.28]{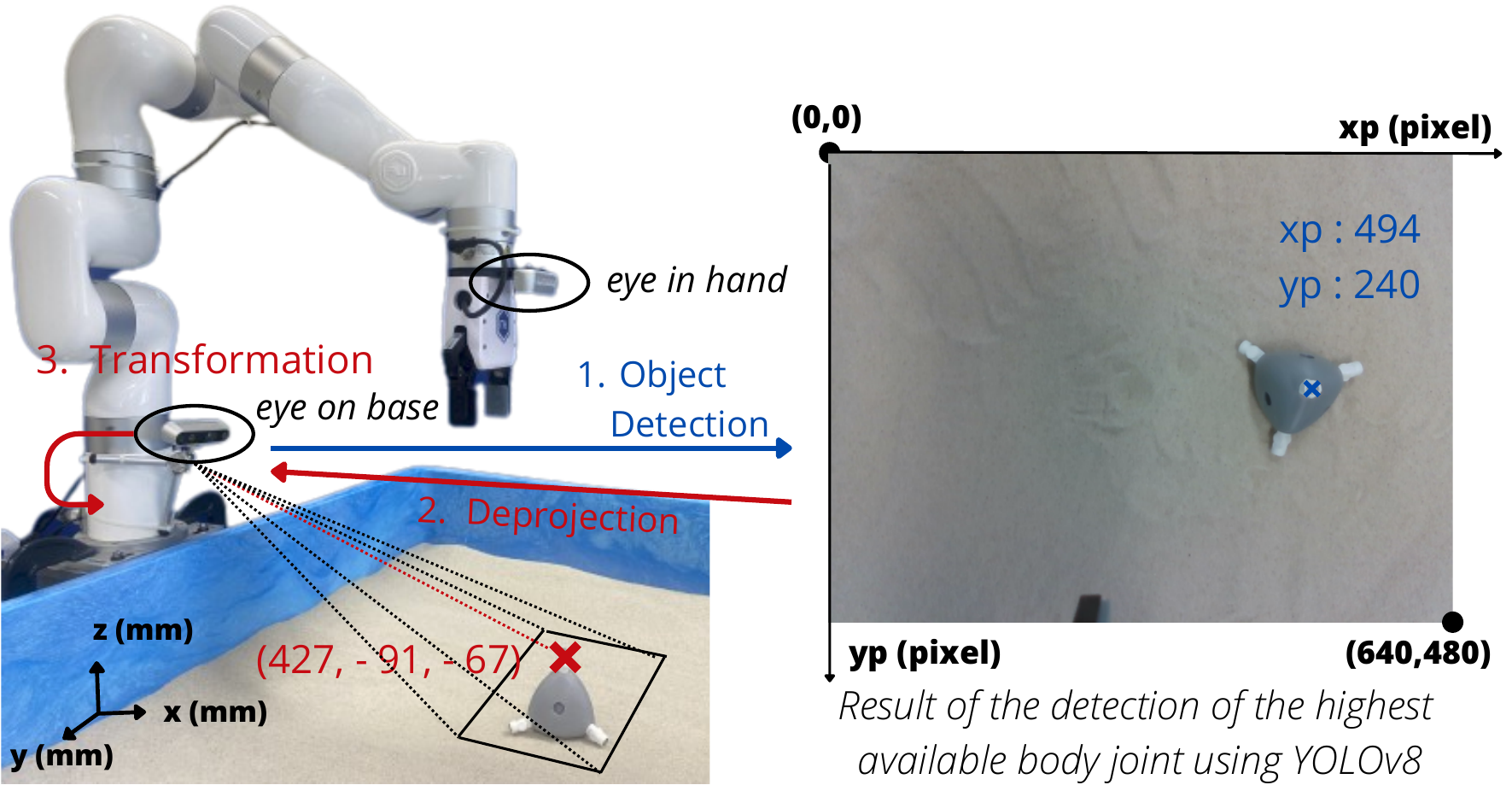}
      \caption{Scheme of the \textit{pixel to point} process.}
      \label{pixel_to_point}
   \end{figure}

In the next, we move from pure detection to real application. Indeed, the instance segmentation gives results in pixels but the xArm moves in the real world. To obtain usable data, we deproject the pixel results to point coordinates in millimeters (mm) using the depth information and the intrinsic parameters of the RealSense camera. The intrinsic matrix K contains the focal lengths and the principal point. We eventually transform these coordinates from the camera frame to the xArm's frame using TF ROS package to retrieve the final coordinates in the robot's workspace \fig{pixel_to_point}.

Now that we can deproject specific points from pixel to coordinates we can grasp the rock. To perform a better grasping, with oriented rocks for instance, we will use a specific package GPD described in the following subsection. 

The final step is to actually stack the rock. After grasping, a new instance segmentation is performed from the \textit{eye on base} camera's frames and the deprojection process can be repeated to calculate the height. The xArm is sent to the determined final position, with an accurate \textit{z} value given the rock height and the depth of the stacking point.



\subsection{Modular robot model assembly task}

The goal of this task is to assemble the prototype of our modular robot model composed by modular components as shown in \fig{custom_datasets}(c). This prototype was selected to test our algorithms to demonstrate the challenging task of grasping. In order to implement the assembly using our general framework integrating GPD and YOLOv8 for tasks planning,  we need to implement the specific sequence and use the specific modules described in \fig{software_architecture}.

\begin{figure}[b]
      \centering
      \includegraphics[scale=0.4]{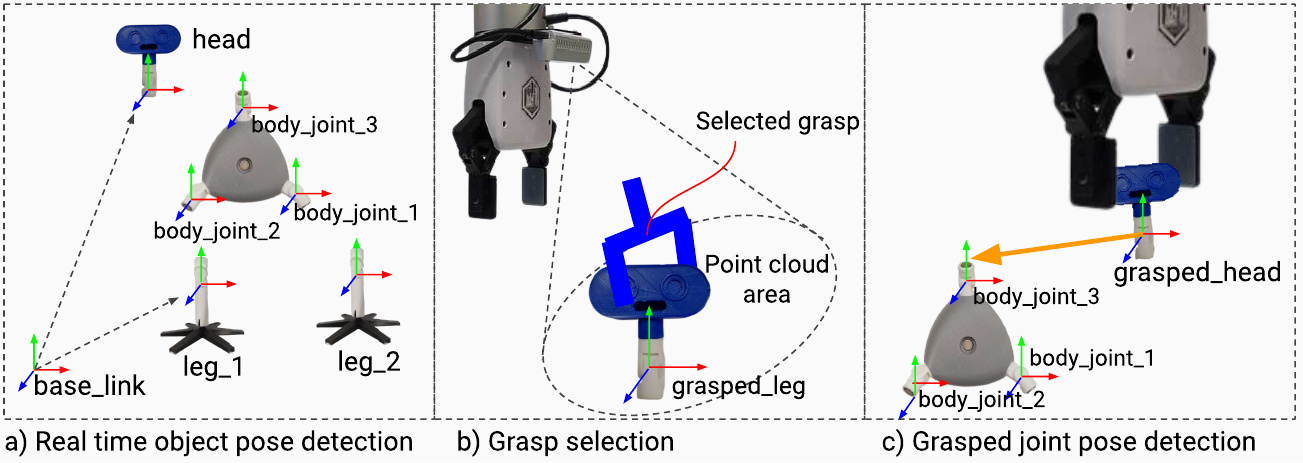}
      \caption{Main steps for the robot assembly task.}
      \label{assembly_steps}
\end{figure}

The main steps for this task are the \textit{Get object pose} that implements the call to the \textit{Object2workspacePose} class to retrieve the object pose in the workspace, \fig{assembly_steps} a). The \textit{Grasping sequence} that moves the \textit{eye in hand} camera close to the object, centering the \textit{pointcloud} on the object and allowing GPD library to generate the grasp candidates around the object of interest, \fig{assembly_steps} b). Once a valid grasp pose has been received and the actual grasping sequence has finished, we move the part to a \textit{pre-assembling} position to detect the joint assembling point position and calculate the displacement to the target body joint and assemble the part, \fig{assembly_steps} c).

\section{Experiments}

\subsection{Rock stacking}

The first step in validating the algorithm is the rock detection. We evaluate the mAP of yolov8m model trained on our \textit{moonrock dataset} Table III. We provide different validation sets by adding augmentation steps such as brightness, exposure and noise. On the non-modified set, the model obtained very good results with 74.0 mAP50-95 for both the box and the mask. On the worst set, we still observed acceptable accuracy with 46.7 and 45.8 mAP50-95. The training dataset is only made with the imitated rocks, so we tested the transposition to real rocks and the model successfully performed 69.5 mAP50-95 on mask segmentation, compared to 77.0 with fake rocks. Finally, we can notice a small decrease in the accuracy of small rocks compared to big ones.

\begin{table}[b]
\caption{YOLOv8 Moonrock model object detection accuracy.}
\label{moonrock_model_validation_table} 
\begin{center}
\begin{tabular}{|c||c||c|}
\hline
Validation Set & mAP50-95 box& mAP50-95 mask\\
\hline
normal  & 74.3 & 74.0\\
\hline
b/e* & 73.7 & 72.7\\
\hline
noise 5 & 46.9 & 46.1\\
\hline
b/e* - noise 5 & 46.7 & 45.8\\
\hline
\hline
small rocks& 73.8 & 72.3\\
\hline
big rocks & 75.0 & 74.9\\
\hline
fake rocks & 76.7 & 76.6\\
\hline
real rocks & 71.1 & 69.5\\
\hline
\end{tabular}
\end{center}
*augmentation step: brightness +/- 25\% and exposure +/- 20\%

\end{table}

To evaluate the algorithm of rock stacking, we perform 50 tests. In the grasping strategy, the rocks need to be sorted by size first. We evaluate the size classification success rate of 96\%; the mask area sorting works very well, even on rocks with only 1 cm of difference in length. The two failures are because of very high exposure on a rock which induces a small error on the mask as show in \fig{custom_datasets}(b).

The use of instance segmentation also results in efficient height estimation, with very good accuracy and only 4\% of relative error.

The rock-stacking task has an overall success rate of 92 \%. The grasping failed twice, and twice the rock is grasped at the extreme end, so when it is stacked, the rocks' center of mass are shifted and the rock topples over. We measured an average alignment error (distance from bottom rock's center to the top one) of 25 mm. To correct this error, we should use the second camera to get a feedback of the grasp and maybe track the rock while it is stacking up.

The rock-stacking task in a moon-like environment faces several challenges. The first is to provide highly accurate results in object detection and instance segmentation in difficult light conditions (strong light variations, shadow occlusions or exposure and brightness); we show that constructing a custom dataset and training the YOLOv8 model on it can overcome these difficulties. Then, we manipulate irregular rocks so we perform instance segmentation to sort the masks' area to get an accurate size classification  and we integrate GPD package in the grasping strategy to autonomously grasp almost any kind of rock . Finally, we tackle the non flat surface challenge, making the rocks' height calculation difficult, by introducing a second camera to perform instance segmentation.

\subsection{Robot assembling}
For this task we aim to assemble the robot model shown in \fig{moon}, that consist of the parts shown in \fig{custom_datasets}c). The success of the assembling depends mainly on three factors, associated to the main steps presented on \fig{assembly_steps}, i) the accuracy and stability of the object pose detection -\tab{pose_accuracy}-, ii) the success of the selected grasp -\tab{grasp_success}- and iii) the visibility of the grasped joint -\tab{grasp_success}-, which depends on the actual grasped orientation. We evaluate these factors separately for the head and the leg objects. For i) we put the objects in a fixed position and recorded the position for 144871 samples, and calculated the standard deviation for every coordinate as shown in \tab{pose_accuracy}. We can see that for the head and legs, the maximum error is 0.37 mm, which is pretty accurate to define the grasp trajectory. However, for the body joints the maximum error is 36.16 mm; this is due to the small size of the joints and the noise in the depth frame used for the de-projection of the pose.

To evaluate the accuracy of the selected grasps, we repeat the grasping sequence from different initial positions of the objects, the results are shown in \tab{grasp_success}. The success in this step is very dependent on the selected grasp, which due to the GPD implementation is inherently stochastic. More fine tuning of the parameters can be done for a specific object but it will affect the performance of other objects.

For the evaluation of the visibility of the grasped joint, after a successful grasp of the object, we send the \textit{eef} to the fixed \textit{pre-assembly} position and calculate the success ratio of pose detection, the results are shown in \tab{grasp_success}.

The main challenge of this task is to be able to detect and manipulate small objects, as well as detecting the target positions for assembling. We approach this problem using a real-time system that allows us to calculate several validity checks to improve the success ratios of the assembly pipeline. Another challenge for this experiment is the non-optimal trajectories generated by MoveIt for some cases, which we solve by planning directly in the joint space for those particular cases. We partially tackle occlusion problems by having two cameras, however, there are still limitations particularly in assembling the legs. We plan to improve this by using a camera on a second arm in the near future. The final critical issue is the limitation of YOLOv8 to provide non-oriented bounding boxes, which is required for a more precise assembly. We plan to do more post-processing of the YOLOv8 results in order to estimate the angle.\\
Even though we use a very simple and small robot model, we achieved high success rates in the assembly process, so for the future assembly tasks of real robots and bigger parts we expect to improve our results.

\begin{table}[t]
\caption{Object pose estimation accuracy (144871 samples).}
\begin{center}
\label{pose_accuracy}
\begin{tabular}{|c|ccc|}
\hline
\multirow{2}{*}{Object} & \multicolumn{3}{c|}{Standard deviation [mm]}    \\ \cline{2-4} 
                        & \multicolumn{1}{c|}{$\sigma_{x}$} & \multicolumn{1}{c|}{$\sigma_{y}$} & $\sigma_{z}$ \\ \hline
        head            & \multicolumn{1}{c|}{0.37}  & \multicolumn{1}{c|}{0.21}  &  0.34 \\ \hline
        leg             & \multicolumn{1}{c|}{0.35}  & \multicolumn{1}{c|}{0.13}  &  0.33 \\ \hline
        body joint L    & \multicolumn{1}{c|}{6.48}  & \multicolumn{1}{c|}{36.16}  &  0.18 \\ \hline
        body joint U    & \multicolumn{1}{c|}{6.99}  & \multicolumn{1}{c|}{19.45}  &  28.31 \\ \hline
\end{tabular}
\end{center}
\end{table}

\begin{table}[t]
\caption{Grasp success and Grasped joint position detection success.}
\label{grasp_success}
\begin{center}
\begin{tabular}{|c|c|c|c|}
    \hline
    Object & Attempts & Grasp success & Grasped joint detection    \\ \hline
    head   & 14    & 92.8\%  & 95.3\% \\ \hline
    leg    & 42    & 76.1\%  & 86.2\%\\ \hline
\end{tabular}
\end{center}
\end{table}

\section{Conclusion}

This paper is the first milestone for the integration of our vision-based systems on robotic manipulators for lunar applications. The explanations of our software framework, its integration for xArm7 and our experiments demonstrate how an integration of the same vision-based software can be used for various robotic applications. The results of these vision-based frameworks can be used in many other ways to improve the performance. For instance, using real-time instance segmentation for tracking the pose of objects for a better manipulation, like stacking or assembling. The next achievement is to autonomously assemble a full-scale modular robot. In addition to the presented software, additional features can be introduced such as a second arm, assembly sequence planning as well as communication between several robots.

\addtolength{\textheight}{-12cm}   




\end{document}